\documentclass[letterpaper, 10 pt, conference]{ieeeconf} 
\usepackage{geometry}
\IEEEoverridecommandlockouts                              
\usepackage{cite}
\usepackage{amsmath,amssymb,amsfonts}
\usepackage{algorithmic}
\usepackage{graphicx}
\usepackage{textcomp}
\usepackage{xcolor}
\usepackage{booktabs}
\usepackage{array}
\def\BibTeX{{\rm B\kern-.05em{\sc i\kern-.025em b}\kern-.08em
    T\kern-.1667em\lower.7ex\hbox{E}\kern-.125emX}}
\newcommand{\eqdef}{\mathrel{\mathop:}=}

\title{\LARGE \bf Contrastive Masked Autoencoders for Character-Level Open-Set Writer Identification}

\author{Xiaowei Jiang$^{1}$, Wenhao Ma$^{1}$, Yiqun Duan$^{1}$, Thomas Do$^{1}$, Chin-Teng Lin$^{1*}$%
\thanks{$^{1}$GrapheneX-UTS Human-centric AI Centre, Australian AI Institute, School of Computer Science, Faculty of Engineering and Information Technology, University of Technology Sydney}%
\thanks{$^{*}$Corresponding author: chin-teng.lin@uts.edu.au}
}

\begin{document}

\maketitle
\thispagestyle{empty}
\pagestyle{empty}

\begin{abstract}

In the realm of digital forensics and document authentication, writer identification plays a crucial role in determining the authors of documents based on handwriting styles. The primary challenge in writer-id is the "open-set scenario", where the goal is accurately recognizing writers unseen during the model training. To overcome this challenge, representation learning is the key. This method can capture unique handwriting features, enabling it to recognize styles not previously encountered during training. Building on this concept,  this paper introduces the Contrastive Masked Auto-Encoders (CMAE) for Character-level Open-Set Writer Identification. We merge Masked Auto-Encoders (MAE) with Contrastive Learning (CL) to simultaneously and respectively capture sequential information and distinguish diverse handwriting styles. Demonstrating its effectiveness, our model achieves state-of-the-art (SOTA) results on the CASIA online handwriting dataset, reaching an impressive precision rate of 89.7$\%$. Our study advances universal writer-id with a sophisticated representation learning approach, contributing substantially to the ever-evolving landscape of digital handwriting analysis, and catering to the demands of an increasingly interconnected world.
\end{abstract}

\section{Introduction}

\begin{figure}[h!]
  \centering
  \includegraphics[width=0.45\textwidth]{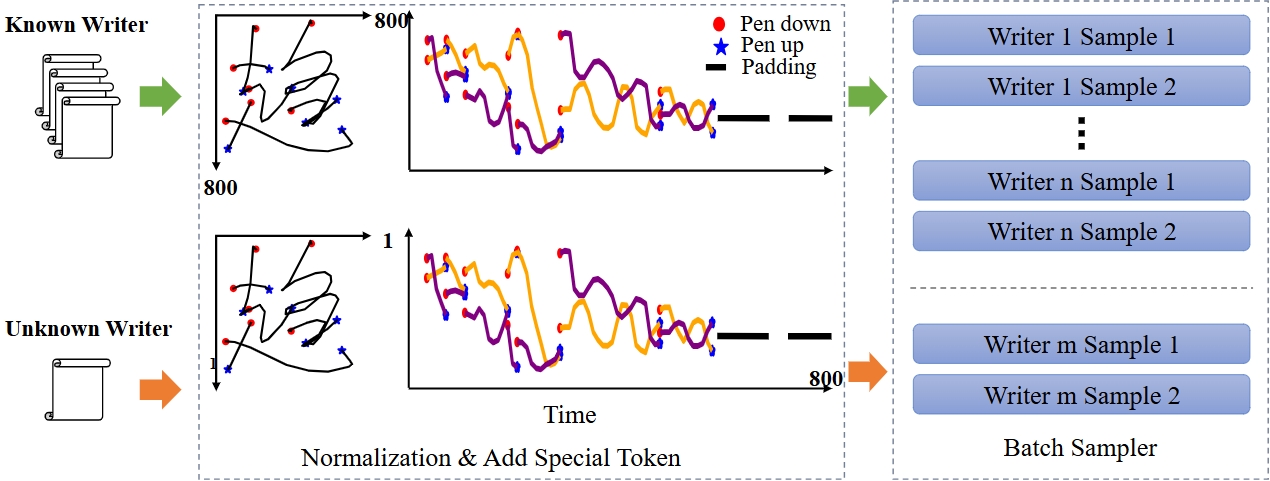}
  \caption{This figure depicts the process for open-set scenarios, highlighting the distinction between the training and testing phases. Training is performed using data from one group, while testing is done with individuals not seen during training. Our method involves training the model by creating pairs of trajectories, \(\tau_{ni}\) and \(\tau_{nj}\), from the same writer \(n\). This technique is intended to enable the model to learn the similarities within each writer's handwriting and the differences between writers. Testing is then carried out on data from unknown writers \(m\). The middle subfigure illustrates the normalization process in our method, which scales the character trajectory to the range (0,1].}
  \label{fig_data_loader}
\end{figure}

In writer identification (writer-id) systems, particularly in forensic sciences, the primary aim is to identify the author of handwritten documents\cite{jain2004introduction}. A major challenge in these systems arises in open-set scenarios(example is shown in Fig \ref{fig_data_loader}), where accurately identifying authors not represented in the model's training data is crucial. Recently, the field of online writer identification has emerged as a more focused area of interest, surpassing traditional methods of digital handwriting analysis \cite{venugopal2017online}. This shift towards analyzing handwriting trajectory data signifies a major development in the field \cite{ghosh2022advances}. However, addressing open-set scenarios in online writer identification tasks remains an unresolved challenge.
This research is significant for its potential to improve security by accurately attributing authorship in legal and forensic cases, even with partial handwriting and without a comprehensive database. Additionally, it has applications in historical document analysis, offering insights into cultural and historical contexts.

Contrasting with online systems, offline systems, denoted as \( \textbf{O} \), capture handwritten documents as images and employ image processing techniques to achieve similar identification goals \cite{hagstrom2022writer,he2021gr,koepf2022writer,he2020frag}. In these systems, the matrix \( \textbf{O} \) can be represented where each element \( p_{ij} \) corresponds to the pixel intensity at the \( i^{th} \) row and \( j^{th} \) column of the image. This matrix captures essential spatial features of handwriting such as stroke thickness and letter shapes, but offers limited temporal information. 

Training robust writer-id models is important but challenging. Generally, writer-id is perceived as a multi-class classification problem, where a writer is identified by comparing handwriting samples against reference \cite{alaei2022handwriting}. Some progress in both offline \cite{he2021gr,koepf2022writer,he2020frag}, \cite{li2023research,mohammad2022self} and online writer-id \cite{venugopal2017online}, \cite{chen2021letter,faundez2021online,zhang2022msds,jiang2022dsdtw,yang2016deepwriterid} tasks has been made. Traditional online writer-id systems, like those by Schlapbach et al., use hand-crafted features such as direction and curvature \cite{schlapbach2008writer}. Recently, Artificial Neural Networks (ANN) have been increasingly used in online writer identification, yet they face certain limitations. Methods by Yang \cite{yang2016deepwriterid}, Venugopal \cite{venugopal2017online}, and Uchida \cite{koepf2022writer} primarily analyze text levels, necessitating lengthy texts. Yang et al. \cite{yang2015chinese} introduced a character-level method, but it was only evaluated in closed-set scenarios. 
Chen and Wu \cite{chen2021letter} developed a letter-level system requiring minimal data, applicable in both closed and open-set scenarios, with a significant disparity in rank-1 accuracy \cite{moon2001computational} across these conditions. 
Their findings indicate a notable achievement, but addressing the open-set scenario at the character level, particularly for single-character tasks, continues to be a challenge.

\begin{figure}[h!]
    \centering
    \includegraphics[width=0.4\textwidth]{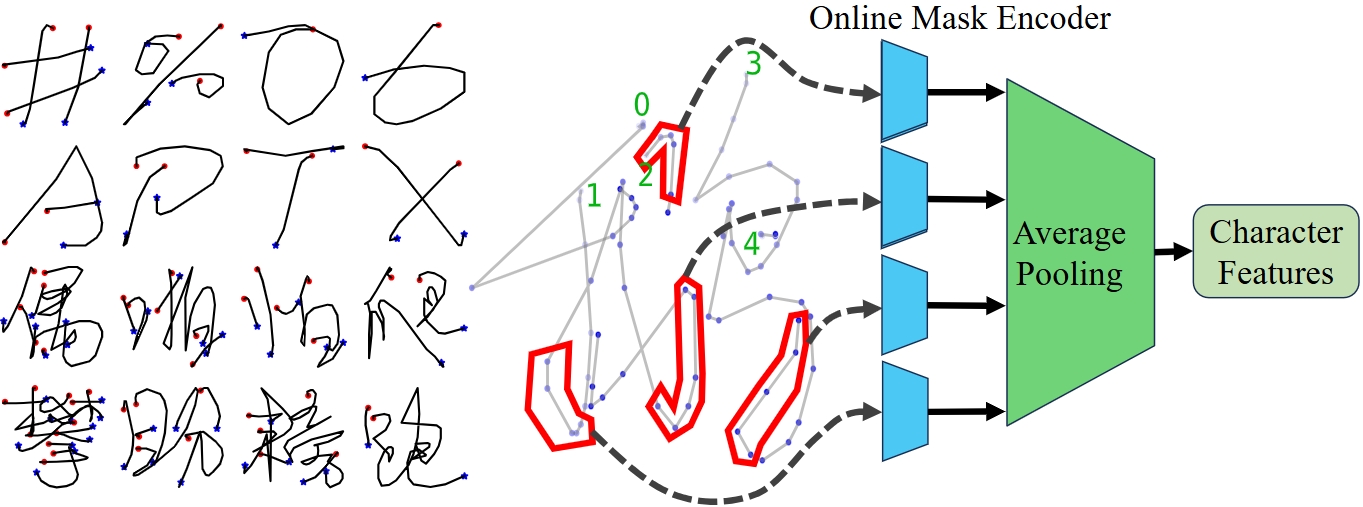}
    \caption{Illustration of the CMAE model's training methodology. \textbf{Left}: Sample images from CASIA-OLHWDB. \textbf{Right}: Depiction of the CMAE encoder's method for extracting features from handwriting character trajectories. The CMAE encoder receives unmasked segmented patches as input, depicted as a red curve in the character "geese", which are randomly chosen and governed by the mask ratio.}
    \label{fig_into}
\end{figure}

Our research is driven by the concept of enhancing writer identification in online single-character open-set scenarios. We focus on developing a system capable of accurately identifying writers, particularly in challenging open-set scenarios where the writer's style may not be present in the training dataset based on only one character. This endeavor is crucial for various applications, from authenticating historical documents to aiding in criminal investigations.

To address this challenge, our CMAE framework combines the strengths of Mask Auto-Encoders (MAE) \cite{he2022masked}  and Contrastive Learning (CL)\cite{oord2019representation}. MAE focuses on intricate feature extraction and reconstruction, which are essential for capturing sequential aspects of handwriting like stroke patterns and speed dynamics. The efficacy of Transformers encoder in the realm of handwriting writer-id has already been established in the field, showcasing its potential \cite{koepf2022writer}, \cite{he2022review,jungo2023character}. The MAE paradigm, leveraging the strengths of the Transformer encoder, brings forth a unique approach to handling handwriting data. This process involves masking parts of the handwriting input and then reconstructing them, thereby enabling the model to understand various handwriting styles deeply. Meanwhile, CL enhances the model's ability to differentiate between styles by comparing handwriting samples. It assesses similarities and differences across samples, sharpening the model's discriminative capabilities. Furthermore, CL has shown considerable promise in the realm of representation learning, particularly in the context of text recognition \cite{luo2022siman}. This integration of MAE and CL allows CMAE to strike a balance between detailed feature analysis and style differentiation, crucial for handling unseen handwriting trajectories with short lengths in open-set scenarios. 

The contribution could be summarized in fourfold:
\begin{itemize}
    \item It pioneers the use of single characters for writer-id tasks.
    \item It introduces an innovative framework tailored for open-set settings. 
    \item It initially integrates MAE with CL to enhance the capacity for representation learning, thereby significantly boosting the accuracy and reliability of the writer-id task.
    \item Our approach achieves a remarkable precision of 89.7\% ± 7.4\% on the CASIA Online Database dataset (CASIA-OLHWDB) dataset (Fig \ref{fig_into}), setting a new standard for state-of-the-art (SOTA) models in the domain of open-set online writer identification.
\end{itemize}

\section{Related Work}

\subsection{Writer Identification (writer-id)}
Writer-id has seen significant breakthroughs in recent years, thanks to numerous studies in the field \cite{chen2021letter}, \cite{jiang2022dsdtw}, \cite{mehralian2023self}. Numerous methods have emerged as efficient ways to identify the authorship of handwritten texts. Some research based on CNN-RNN \cite{he2021gr}, CNN \cite{yang2015chinese}, RNN \cite{zhang2017end} and LSTM \cite{chen2021letter}/BLSTM or weighted majority vote row-decision model \cite{cilia2020end} are typical writer-id methods. These approaches treat the handwriting documents or texts as images, extracting the visual features like colors, textures, and shapes and learning these visual representations for identification. Especially, Vision Transformer(ViT) \cite{dosovitskiy2021image}, \cite{hassani2021escaping} models project the flattened patches into an embedding space with multi-head attention layers, which work well in different scripts (Latin and Greek) and different writing styles (cursive handwriting and block letters) \cite{koepf2022writer}. In the Online writer-id field, Dynamic Time Warping (DTW) \cite{sakoe1978dynamic} is one a popular technique for sequence alignment and Jiajia et al. \cite{jiang2022dsdtw} designed a DTW-based signature verification system. Chen et al. \cite{chen2021letter} indicated a hierarchical attention pooling (HAP)-based model to fuse letters with multiple writing styles into a compact feature vector, and Christlein et al. \cite{christlein2018handwriting} first apply self-supervised representation learning method in offline data, and Pouya et al. \cite{mehralian2023self} indicated separating the trajectory of handwriting character into sub-windows can improve the representation learning of handwriting trajectory. More details and advances about the writer-id task can be seen in reviews \cite{ghosh2022advances}, \cite{alaei2022handwriting}.

\subsection{Open-Set Settings}
Open sets, or systems without enrollment challenges, demand the identification of subtle and unique authorial cues in texts, even when these specific samples or styles have not been encountered in the training set. To discern such nuanced authorship cues in open-set identification tasks, models require enhanced capabilities in feature extraction and pattern generalization. Currently, just a handful of online systems address the open-set problem, for example, DTW-SVM \cite{gargouri2013text}, biLSTM-based \cite{chen2021letter}, and DsDTW \cite{jiang2022dsdtw} proposed some solutions. As for offline systems, Christlein et al. \cite{Christlein2015} utilize local features to train a CNN with image patches centered on handwriting. Conversely, Tang and Wu \cite{Tang2016} focus on global features, processing images with multiple handwriting lines in one pass to simplify the encoding process. Keglevic et al. \cite{Keglevic2018} propose a Triplet CNN and Liang et al. \cite{Liang2021} apply transfer learning based on a pre-trained ResNet-50 \cite{He2016}. Except to CNN, ViT-Lite-based \cite{koepf2022writer} approaches are also applied.

\subsection{Representation Learning}
Representation Learning \cite{Bengio2013} refers to the extraction of meaningful patterns from raw data by machine learning algorithms to create simpler and more comprehensible representations. CL has emerged as a key paradigm, focusing on enhancing the similarities between comparable (positive) samples and diminishing those among different (negative) samples \cite{oord2019representation}. Seminal works \cite{Chen2020Big, Chen2020Simple} like SimCLR has advanced this domain, notably through large-batch training methods and the use of negative samples. In text recognition, models such as CMT-Co by Zhang et al. \cite{Zhang2022} and SCLAiR by Tripathi et al. \cite{Tripathi2022} have demonstrated the efficacy of supervised CL in text representation, while self-supervised CL has shown promise in handwriting analysis, especially in digital paleography identification \cite{Lastilla2022} and text recognition \cite{Guan2023}. Another significant stride in this field is the development of MAE \cite{he2022masked}, \cite{Chen2023Masked, Chen2023Context} a self-supervised approach emphasizing image reconstruction, adept at capturing diverse data features essential for comprehensive data understanding. This method employs masking for reconstruction, enhancing the encoder to learn intricate relationships within data.

\section{Methodology}
In this section, we will explain the design of CMAE, a novel representation learning framework to identify the writers based on single characters. We first introduce the dataset preparation. Then, we explain the CMAE in detail. Finally, we provide the training and evaluation details.

\subsection{Task Definition}

The writer-id problem can be formally expressed as follows: Given a dataset \( \mathcal{D} \) consisting of handwriting samples with $N$ writers, where writers donate pairs of handwriting trajectories with sample pairs ($\tau_{i}$ and $\tau_{j}$). the goal is to learn a function $f: (\tau_{i} \text{ and } \tau_{j}) \rightarrow \sigma$ that maps two handwriting trajectories to their writer's identity probability distribution $\sigma$.

\begin{figure*}[ht]
\centering
\includegraphics[width=0.65\textwidth]{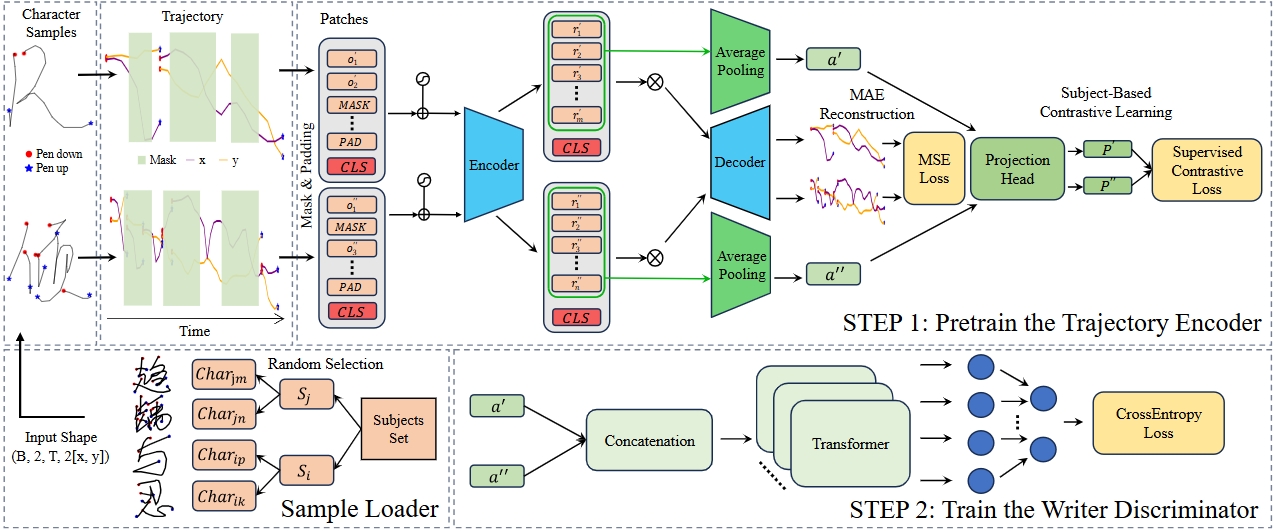}
\caption{Schematic of Our Proposed Model. \textbf{STEP1}: This framework starts by processing point sequences of single characters (including English, Chinese, and symbols). For instance, the trajectory for "R" (depicted at the top in the figure) consists of coordinates \((x_i,y_i)\) at each time \(t_i\), along with special tokens for pen down, pen up, and padding (not shown in this figure). The "R" is then segmented into patches and subjected to random masking. Subsequently, it passes through a transformer-based encoder, which extracts and maps the trajectory features of "R" into an embedding space. These features are then directed into two distinct modules: the CL and the MAE. The MAE decoder reconstructs the masked patches using the unmasked ones, while the CL adjusts the distance between two embeddings \((a^{\prime}, a^{\prime\prime})\) of trajectories, either from the same or different individuals. \textbf{STEP2}: The embedding \(a^{\prime}\) and \(a^{\prime\prime}\) are then processed through a transformer block followed by an MLP, culminating in the computation of CrossEntropy Loss.}
\label{fig_model_arch}
\end{figure*}

\subsection{Framework}
\subsubsection{CMAE Encoder}

The architecture of CMAE, as shown in Fig \ref{fig_model_arch}, inspired by MAE, incorporates a Transformer-based encoder and decoder, with a novel twist in its design that allows for the processing of input data segmented into patches and includes an innovative approach to mask these patches selectively. Specifically, the trajectory data $tau_n$ is patched into segments $\textbf{P}\eqdef\{p_i\}^N_{i=1}$. Then, we mask patches derived from the unpadded segments of trajectory data as $\textbf{P}^m$, since masking padded areas would not contribute meaningfully to our model's learning process. This modification is crucial due to the varying lengths of each sample's mask, making it infeasible to maintain them as a tensor in the original MAE design. The unmasked patches ($\textbf{P}^v$) are fed into the Transformer encoder $f_{enc}(\cdot)$ and get the embedding features $\textbf{Z}^v\eqdef\{\textbf{Z}^v_i\}^N_{i=1}$ as:

\begin{equation}
\textbf{Z}^v = f_{enc}(\textbf{P}^v)
\end{equation}

Our encoder consists of eight Transformer blocks, each with eight attention-heads. The input trajectory size is set at 800×2, with patch sizes at $5×2$. We project the flattened patches (n\_patches = 160) into an embedding space with a dimension of 512, utilizing multi-head attention mechanisms. 

\subsubsection{CMAE Reconstruction Decoder}


After being encoded by the encoder, the embedding features $\textbf{Z}_v$ of unmasked patches together with masked patches feature vector $\textbf{Z}_m$ are fed into the feature decoder to reconstruct the unmasked patches $\textbf{P}^m$. The decoder $f_{dec}(\cdot)$ is also transformer-based architecture.

\begin{equation}
\textbf{P}^m = \mathbb{I} \cdot f_{dec}(\textbf{Z}^v, \textbf{Z}^m)
\end{equation}

where $\mathbb{I}$ denotes an indicator that only selects the masked feature vectors. We have also refined the reconstruction loss calculation to focus exclusively on masked and unpadded patches. The reconstruction decoder operates in a 256-dimensional space and mirrors the encoder's structure with eight Transformer blocks and eight attention-heads.

\subsubsection{CMAE Contrastive Learning}
For the CL module, we employ average pooling on all unmasked embedding post-encoding. This is followed by a two-layer Multi-Layer Perception (MLP) with ReLU activation, which serves as a projector $f_{proj}(\cdot)$ for calculating the cosine similarity $\cos \theta$ of two trajectory features $z^v_i$ and $z^v_j$ as:

\begin{equation}
\cos \theta = \frac{\mathbf{A} \cdot \mathbf{B}}{\|\mathbf{A}\| \|\mathbf{B}\|}
\end{equation}

where $A = f_{proj}(z^v_i)$ and $B = f_{proj}(z^v_j)$, forming the basis of our CL loss \cite{Khosla2021}. We define trajectories from the same individual as positive pairs and those from different individuals as negative pairs. 

\subsubsection{CMAE Discriminator}
The architecture of the Writing Discriminator $f_{dis}(\cdot)$ consists of multiple stacked transformer blocks, followed by an MLP block, and then a softmax layer for outputting the prediction distribution. The rationale for employing transformers in this design is their capability to capture relationships of trajectory features between sample pairs. This structure leverages the transformers' strength in handling sequential data and their efficiency in discerning complex relational patterns within sequences. The input to this Writing Discriminator is the concatenation $\textit{concat}(\cdot, \cdot)$ of paired trajectory feature embedding $z_i, z_j$, which is the output of the CMAE's encoder. 

\begin{equation}
\sigma(\mathbf{z})_{i,j} = \text{SoftMax}(f_{dis}(\textit{concat}(z_i, z_j))
\end{equation}
Here, the $\sigma(\mathbf{z})_{i,j}$ represents the softmax distributions with pair ($z_i, z_j$).

\subsection{Training Objective}
Our model employs a combination of three distinct loss functions to optimize its performance:
\textbf{Reconstruction Loss.} We utilize Mean Squared Error (MSE) Loss to minimize the disparity between reconstructed trajectories and the original ones, ensuring high-fidelity restoration of the original data.
\textbf{Contrastive Learning Loss.} For the contrastive learning process, we leverage the Supervised CL Loss \cite{Tripathi2022} to optimize the CMAE encoder. This loss enhances the discriminative capacity of the model by learning from positive (same individual's samples) and negative pairs (different individuals' samples).
\textbf{Cross-Entropy Loss.} The CMAE Discriminator employs Cross-Entropy Loss ($\ell_{CE}$) to determine whether two trajectories originate from the same individual.
\textbf{Loss Integration.} To effectively merge the Reconstruction (RE) and Contrastive Learning (CL) losses, we use a weighting strategy $\lambda \in (0, 1)$ to balance their contributions, ensuring a balanced optimization of both aspects.
\begin{equation}
\ell = \lambda\ell_{\text{RE}}+(1-\lambda)\ell_{\text{CL}}
\label{all_loss}
\end{equation}

\section{Experiment}
In our experiments, we primarily focus on the open-set scenario to evaluate the effectiveness of our results. By testing in an open-set context, we aim to provide a more comprehensive and realistic evaluation of the model's performance, particularly in its ability to generalize and accurately process data beyond the distribution of the training dataset.

\subsection{Dataset Preparation}\label{AA}
\subsubsection{Dataset Description}
We describe the details of the two databases we have selected as follows.

\textbf{CASIA Online Database dataset (CASIA-OLHWDB)}
We utilize the CASIA Online Database dataset (CASIA-OLHWDB1.0-1.2) \cite{liu2011casia} for both training and testing our model. Fig \ref{fig_into} shows some samples from this dataset. This comprehensive dataset comprises approximately 3.9 million samples across 7,356 classes, which include 7,185 Chinese characters and 171 symbols, contributed by 1,020 individuals. The dataset is partitioned into distinct training and testing sets, ensuring no overlap of individuals between the two. The dataset is organized into three versions, DB1.0, DB1.1, and DB1.2, with distinct distributions of individuals. In DB1.0, the dataset is split into 336 individuals allocated for training and 84 for testing. For DB1.1, the distribution includes 240 individuals for training and 60 for testing. Similarly, DB1.2 follows the same pattern as DB1.1, with 240 individuals designated for training and 60 for testing. In total, these results in 816 individuals for training and 204 for testing. Further details about the dataset can be found in the reference paper \cite{liu2011casia}. For the model training process, we split the training dataset (n = 816) into training and validation subsets in an 8:2 ratio, resulting in 653 individuals for training and 163 for validation.

\textbf{IAM On-Line Handwriting Database (IAM-OnDB)}
This database, focused on English handwriting samples, originally contained contributions from 221 writers. However, due to some samples lacking identified writer IDs, we only focus on labeled pages, narrowing our dataset to handwriting from 197 writers. The IAM-OnDB is characterized by its collection of 13,049 isolated and labeled text lines in an online format. These lines are constructed from a total of 86,272 words, which are drawn from an extensive vocabulary pool of 11,059 distinct words. We tailor the dataset to suit our model's needs and implemented a strategy of randomly cropping line data to a maximum length of 800 timepoints. For the training phase, we divide the pool of 197 writers into training and validation subsets by an 8:2 ratio. Consequently, this results in a distribution of 158 individuals for training and 39 for validation.

\begin{table}[h]
\centering
\caption{Comparison with other SOTA models under the open-set setting in IAM-OnDB Dataset}
\label{tab:iam}
\begin{tabular}{>{\arraybackslash}p{0.55\columnwidth}c}
\toprule
\textbf{Methods} & \textbf{Rank-1 (\%)} $\uparrow$ \\
\midrule
Point based(Gargouri et al. 2013)  & 9.7 \\
Histogram based(Dwivedi 2016)  & 7.4 \\
DeepWriterID(Yang et al. 2016)  & 52.5 \\
DeepRNN(Zhang et al. 2017)  & 57.4 \\
Multi-branch encoder(Chen et al. 2021) & 76.2 \\
\textbf{CMAE (Ours)}  & \textbf{81.6} \\
\bottomrule
\end{tabular}
\end{table}

\subsubsection{Data Preprocessing}
Two datasets follow the same data preprocessing approach.


\textbf{Special Tokens Introduction.} To address specific states in the handwriting process, we introduce three special tokens: pen-up (-0.01, 0), pen-down (-0.01, -0.01), and padding (0, -0.01). These tokens are carefully chosen to avoid distribution shifting, given that the current data values range from [0,1).

\textbf{Padding.} All data are uniformly padded to a length of 800 timepoints to ensure consistency across samples.

\subsection{Model Evaluation}

\subsubsection{Evaluation of Open-Set Writer Identification Setting}

To compare with other previous SOTA models, we utilize the IAM-OnDB to evaluate our model. Given that our original method's implementation strategy differs from previous research, we primarily follow Chen's approach \cite{chen2021letter} for evaluation in IAM-OnDB. In this open-set setting, 20 writers are chosen per selection, and we compute the average rank-1 accuracy \cite{moon2001computational} over 100 random selections in the validation set. Our model achieves an accuracy of 81.6\%±1.6\% (see Table \ref{tab:iam}) in the IAM-OnDB dataset. The IAM-OnDB dataset's small size and the superior performance of representation learning in larger datasets prompted us to use the CASIA-OLHWDB dataset as our primary dataset for further evaluation in the Open-Set Writer Identification Setting. For this, we randomly select 20 writers and 2 characters per writer from the test dataset for each selection. Considering the unpredictability of writer numbers in real-world open-set scenarios, using rank-k accuracy is not a suitable measure for this model design. Hence, to more effectively evaluate our model, we assess the Classifier's performance based on the average accuracy and precision calculated over 100 random selections. The results indicate an average accuracy of 98.1\%±1.4\% and an average precision of 76.0\%±13.9\%.

\subsubsection{Effect of the mask ratio}
We first conducted experiments to investigate the effect of the mask ratio in CMAE. The mask ratio defines the proportion of input data that is masked or hidden for the autoencoder during the training phase. This approach is designed to enhance the model's ability to reconstruct missing or corrupted data, thereby improving its representation learning capabilities.
In our specific context, we hypothesized that varying the mask ratio would significantly impact the model's performance in learning complex data representations. We selected two distinct mask ratios, 0.75 and 0.15, to evaluate this hypothesis. By training two separate models with these mask ratios, we aimed to compare their effects on the model's precision. The results of these experiments are crucial for understanding the optimal balance between the amount of hidden data and the model's performance. This balance is pivotal as it directly influences the model's performance in tasks such as unsupervised feature learning, data denoising, and representation robustness. 
Results in Table \ref{tab:ablation_experiment_rlp} show that when the mask ratio $=$ 0.15, the result (the average accuracy is 99.3\%±0.5\%, and average precision is 89.7\%±7.4\%) is better than mask ratio $=$ 0.75 (the average accuracy is 98.1\%±1.4\%, and average precision is 76.0\%±13.9\%).





\begin{table}[ht]
\centering
\caption{Ablation experiment: the representation learning pretraining processing and mask ratio}
\label{tab:ablation_experiment_rlp}
\setlength{\tabcolsep}{2.3pt} 
\renewcommand{\arraystretch}{1} 
\begin{tabular}{lcccccc}
\toprule
\textbf{Methods} & \textbf{Mask Ratio}&\textbf{Acc(std)$\uparrow$}&\textbf{$\Delta$} & \textbf{Precision(std)$\uparrow$}&\textbf{$\Delta$}\\
\midrule
CMAE & 0.75 & 98.1 (1.4) &/& 76.0 (13.9)&/ \\
CMAE w/o RLP & 0.75 & 75.0 (3.2) &-23.1& 14.3 (2.4)&-61.7 \\
CMAE & 0.15 & \textbf{99.3 (0.5)} &/& \textbf{89.7 (7.4)}&/ \\
CMAE w/o RLP & 0.15 & 91.9 (2.4) &-7.4& 38.8 (7.4)&-50.9 \\
\bottomrule
\end{tabular}
\smallskip
\small \textit{Note: RLP stands for Representation Learning Pretraining.}
\end{table}

\subsubsection{Effect of the representation learning pretraining processing}
In our study, we conducted a series of experiments to examine the influence of pretraining processing on the CMAE Encoder. Pretraining is a pivotal step in neural network development, setting a foundational knowledge base that can significantly impact the model's learning efficiency and performance. For the CMAE Encoder, pretraining involved using a subset of data or a related task to condition the model, hypothesizing that this would enhance its ability to process and learn from complex datasets in the subsequent training phase. The experiments aimed to determine if the optimal pretraining conditions would improve the model's performance. The results show that without the representation learning pretraining processing, the averaged precision reduces to 38.8\%±7.39\% with mask-ratio 0.15. Other results are shown in Table \ref{tab:ablation_experiment_rlp}.

\begin{figure}[h!]
  \centering
  \includegraphics[width=0.2\textwidth]{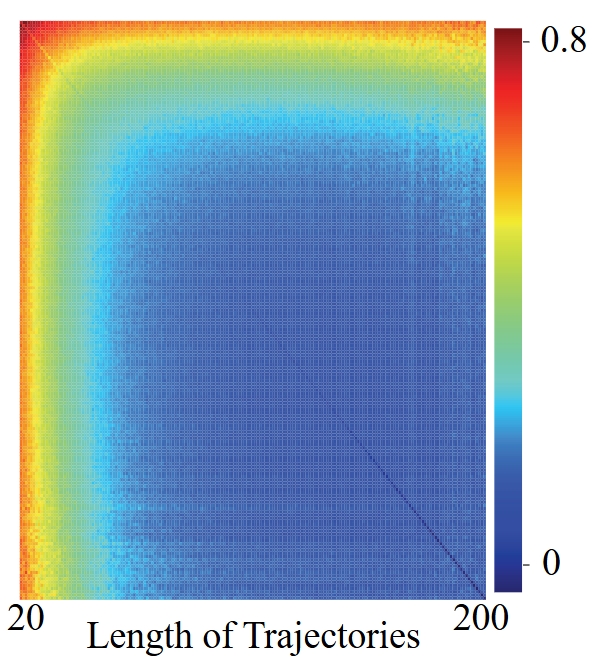}
  \caption{A heatmap illustrating the differences in distance between two handwriting trajectories of varying lengths from the same person. The heatmap's x and y axes range from 20 to 200 timepoints, representing the lengths of the handwriting trajectories. The color scale on the heatmap indicates the average Euclidean distance between two embeddings of these trajectories, derived from the same individual, using the CMAE encoder.}
  \label{fig_length_trajectories}   
\end{figure}

\subsubsection{Effect of Length of Trajectories}
In our study, the length of handwriting trajectories significantly impacts model performance, primarily due to the varying amount of information contained within trajectories of different lengths. To evaluate this effect, we calculated the Euclidean distances between trajectories of various lengths, ranging from 20 to 200 time points, from the same individual. Using our model encoder, we obtained embeddings of these trajectories through the CL projectors and then computed the distances between these embeddings.
Our findings indicate that as the trajectory length increases, the distance between different trajectories decreases, as shown in Fig \ref{fig_length_trajectories}. Notably, when the trajectory length exceeds 40 points, this trend of decreasing distance becomes less pronounced, stabilizing at a distance close to 0.1. This stabilization suggests that representations of longer trajectories from the same individual bear a higher similarity, thereby enhancing the model's ability to accurately identify and distinguish handwriting styles of the same individual. For longer trajectories ($\text{length} > 150$), the diagonal values tend to be close to 0 since most writers have only one trajectory for selection, which means the two trajectories in a pair of samples are the same. These insights not only highlight the importance of trajectory length as a determinant of model performance but also provide crucial guidance for optimizing handwriting recognition models.


\section{Ablation Study}
In our study, we delve into the distinct roles of CL and mask reconstruction in our representation learning framework. To clarify the distinct contributions of each module, we performed ablation studies where we systematically omitted each element to observe its effect on the overall performance of the model. CL, known for its exceptional ability to learn robust and discriminative features by contrasting positive and negative samples, plays a crucial role in enhancing the model's ability to differentiate between varied handwriting styles. On the other hand, mask reconstruction focuses on the predictive reconstruction of masked segments of data, thereby fostering a deeper understanding of the inherent structure and nuances of handwriting. By conducting these ablation studies, we aim to analyze the individual and combined effects of these two modules. This approach will provide us with insightful revelations about how each module contributes to the model’s overall ability to accurately discriminate and differentiate between handwriting samples, enhancing our understanding of the collaboration in representation learning mechanisms. The results presented in Table \ref{tab:ablation_experiment} illustrate that the CL module plays a pivotal role in our model's performance, while the contribution of the MAE is comparatively modest.

\begin{table}[ht]
\centering
\caption{Ablation experiment: representation learning modules}
\label{tab:ablation_experiment}
\begin{tabular}{lcccc}
\toprule
\textbf{Methods}&\textbf{Acc.(std.)$\uparrow$}&\textbf{$\Delta$} &\textbf{Precision(std)$\uparrow$}&\textbf{$\Delta$}\\
\midrule
CMAE & 99.3 (0.5) & / & 89.7 (7.4) & /\\
CMAE w/o CL & 52.0 (3.2)& -47.3 & 6.5 (1.0) & -83.2 \\
CMAE w/o MAE & 98.9 (1.1)& -0.4 & 85.4 (13.1) & -4.3\\
\bottomrule
\end{tabular}
\end{table}

In our study, we also delved into the influence of the depth of transformer-based encoders and decoders on the effectiveness of representation learning. Our comprehensive analysis, as depicted in Table \ref{depth}, clearly demonstrates that increasing the depth of these transformer components directly correlates with improved performance. This finding suggests that models with greater depth are more adept at capturing intricate patterns and features within the data, thereby enhancing their learning and representational capabilities.

\begin{table}[ht]
\centering
\label{depth}
\caption{Ablation experiment: Encoder Depth}
\label{tab:ablation_experiment}
\begin{tabular}{lccc}
\toprule
\textbf{Methods}&\textbf{Depth}&\textbf{Acc.(std.)$\uparrow$}&\textbf{Precision(std)$\uparrow$}\\
\midrule
CMAE & 2 & 80.0 (3.7) & 16.8 (3.6) \\

CMAE & 4 & 86.5 (1.4) & 58.2 (4.3) \\

CMAE & 8 & 99.3 (0.5) & 89.7 (7.4) \\
\bottomrule
\end{tabular}
\end{table}
\section{Conclusion}
This paper presents a character-level open-set writer-id model, CMAE, based on MAE and CL. Our method not only enhances the representational capacity of the model but also significantly improves its ability to discriminate between different writers' styles. Our method demonstrates superior performance over existing models in handling open-set conditions with only single-character handwriting. The experimental results not only demonstrate our model's superior performance over existing models in open-set conditions using single-character handwriting but also highlight its robustness and adaptability in handling unknown or novel data, crucial for real-world applications. This work, therefore, stands as a robust and the SOTA solution to a longstanding problem in digital forensics and document authentication, paving the way for more accurate and reliable writer-id systems in practical, real-world applications.

\textbf{Acknowledgments:} This work was supported in part by the Australian Research Council (ARC) under discovery grants DP180100656 and DP210101093. Research was also sponsored in part by the Australia Defense Innovation Hub under Contract No. P18-650825, US Office of Naval Research Global under Cooperative Agreement Number ONRG - NICOP - N62909-19-1-2058, and AFOSR – DST Australian Autonomy Initiative agreement ID10134. We also thank the NSW Defense Innovation Network and NSW State Government of Australia for financial support in part of this research through grants DINPP2019 S1-03/09 and PP21-22.03.02.

\bibliographystyle{IEEEtran}
\bibliography{main}
\end{document}